\documentclass{article}
\usepackage{ijcai23}

\usepackage{times}
\usepackage{soul}
\usepackage{url}
\usepackage[hidelinks]{hyperref}

\usepackage[small]{caption}
\usepackage{graphicx}
\usepackage{multirow}
\usepackage{booktabs}
\usepackage{tabularx}
\usepackage[switch]{lineno}

\newcommand{\methodname}{\textsc{FedOBD}}

\usepackage[ruled,linesnumbered]{algorithm2e}
\usepackage{mathtools}
\usepackage{amsfonts}
\usepackage{siunitx}
\usepackage{amsmath}
\usepackage{amssymb}
\usepackage{amssymb,amsmath,amsfonts}
\usepackage{ifluatex}
\ifluatex
  \usepackage{unicode-math}
  
\else
  \DeclareMathAlphabet{\mathpzc}{T1}{pzc}{m}{it}

\fi

\DeclareMathOperator*{\argmin}{arg\,min}
\DeclareMathOperator{\sgn}{sgn}

\providecommand{\vect}[1]{\boldsymbol{#1}}

\providecommand{\funname}[1]{\ensuremath{\operatorname{#1}}}
\providecommand{\abs}[1]{\ensuremath{\lvert#1\rvert}}
\providecommand{\ltwonorm}[1]{\ensuremath{{\lVert#1\rVert}_2}}
\providecommand{\infinitynorm}[1]{\ensuremath{{\lVert#1\rVert}_{\infty}}}
\providecommand{\norm}[1]{\ltwonorm{#1}}
\providecommand{\realset}{\mathbb{R}}
\providecommand{\naturalnumberset}{\mathbb{N}}
\providecommand{\defeq}{\ensuremath{\coloneq}}

\providecommand{\subjectto}{\ensuremath{\text{s.t. }}}

\usepackage{cascadia-code}
\usepackage[T1]{fontenc}
\providecommand{\codename}[1]{\normalfont\fontseries{el}\selectfont \texttt{#1}}

\usepackage{xifthen}

\providecommand{\hypothesisspace}{\ensuremath{\mathcal{H}}}

\providecommand{\param}[1][]{\ensuremath{\ifthenelse{\isempty{#1}} {\boldsymbol{\theta}} {\boldsymbol{\theta}_{#1}}}}
\providecommand{\optimalparam}[1][]{\ensuremath{\ifthenelse{\isempty{#1}} {\boldsymbol{\theta}^*} {\boldsymbol{\theta}^*_{\text{#1}}}}}
\providecommand{\finalparam}[1][]{\ensuremath{\ifthenelse{\isempty{#1}} {\hat{\boldsymbol{\theta}}} {\hat{\boldsymbol{\theta}}_{\text{#1}}}}}

\providecommand{\trainingdataset}[1][]{\ensuremath{
\ifthenelse{\isempty{#1}}
{\mathcal{D}_{\text{train}}}
{\mathcal{D}_{\text{train,{#1}}}
}}}
\providecommand{\batch}[1][]{\ensuremath{
    \ifthenelse{\isempty{#1}}
    {\mathcal{B}}
    {\mathcal{B}_{#1}
    }}}
\providecommand{\trainingdatasetsize}[1][]{\ensuremath{
    \ifthenelse{\isempty{#1}}
    {\abs{\trainingdataset}}
    {\abs{\trainingdataset[#1]}}
  }}
\providecommand{\testdatasetsize}[1][]{\ensuremath{
    \ifthenelse{\isempty{#1}}
    {\abs{\testdataset}}
    {\abs{\testdataset[#1]}}
  }}
\providecommand{\batchsize}[1][]{\ensuremath{
    \ifthenelse{\isempty{#1}}
    {\abs{\batch}}
    {\abs{\batch[#1]}}
  }}
\providecommand{\testdataset}{\ensuremath{\mathcal{D}_{\text{test}}}}

\providecommand{\dataitem}{\ensuremath{\boldsymbol{z}}}
\providecommand{\trainingitem}[1][]{\ensuremath{\ifthenelse{\isempty{#1}} {\dataitem_\text{train}} {\dataitem_{#1}}}}
\providecommand{\testitem}[1][]{\ensuremath{\ifthenelse{\isempty{#1}} {\dataitem_\text{test}} {\dataitem_{#1}}}}
\providecommand{\sampleweight}[1][]{\ensuremath{\ifthenelse{\isempty{#1}} {\epsilon} {\epsilon_{#1}}}}

\providecommand{\mlfun}[1][]{\ensuremath{\ifthenelse{\isempty{#1}} {f_{\param}} {f_{#1}} }}

\providecommand{\learningrate}[1][]{\ensuremath{\ifthenelse{\isempty{#1}} {\eta} {\eta_{#1}}}}

\providecommand{\empiricalrisk}[1][]{\ensuremath{\ifthenelse{\isempty{#1}} {\mathcal{R}} {\mathcal{R}_\text{#1}}}}

\providecommand{\fungradient}[2]{\ensuremath{ \nabla_{#2} {#1}}}
\providecommand{\ergradient}[1][]{\ensuremath{ \ifthenelse{\isempty{#1}} {\fungradient{\empiricalrisk}{\param}} {\fungradient{\empiricalrisk}{#1}}}}

\providecommand{\mom}[1][]{\ensuremath{\ifthenelse{\isempty{#1}} {\vect{v}} {\vect{v}_{#1}}}}

\providecommand{\trainingloss}[1][]{\ensuremath{\ifthenelse{\isempty{#1}}
{\mathcal{L}_{\text{train}}}
{\mathcal{L}_{\text{#1}}}
}
}
\providecommand{\validationloss}[1][]{\ensuremath{\ifthenelse{\isempty{#1}}
{\mathcal{L}_{\text{val}}}
{\mathcal{L}_{\text{#1}}}
}
}
\providecommand{\batchloss}[1][]{\ensuremath{\ifthenelse{\isempty{#1}}
{\mathcal{L}_{\batch}}
{\mathcal{L}_{\batch[#1]}}
}}

\providecommand{\batchempiricalrisk}[1][]{\ensuremath{\ifthenelse{\isempty{#1}}
{\mathcal{R}_{\batch}}
{\mathcal{R}_{\batch[#1]}}
}}
\providecommand{\traininggradient}[1][]{\ensuremath{ \ifthenelse{\isempty{#1}} {\fungradient{\trainingloss}{\param}} {\fungradient{\trainingloss}{#1}}}}

\providecommand{\funhessian}[2]{\ensuremath{ \nabla^2_{#2} {#1}}}
\providecommand{\traininglosshessian}[1][]{\ensuremath{ \ifthenelse{\isempty{#1}} {\funhessian{\trainingloss}{\param}} {\funhessian{\trainingloss}{#1}}}}
\providecommand{\empiricalriskhessian}[1][]{\ensuremath{ \ifthenelse{\isempty{#1}} {\funhessian{\empiricalrisk}{\param}} {\funhessian{\empiricalrisk}{\param_{#1}}}}}

\renewcommand{\param}[1][]{\ensuremath{\ifthenelse{\isempty{#1}} {\vect{w}} {\vect{w}_{#1}}}}
\providecommand{\dropoutrate}{\lambda}
\providecommand{\relativeweight}{\beta}
\providecommand{\quantlevelnumber}{s}
\providecommand{\normvar}{d}

\title{\methodname{}: Opportunistic Block Dropout for Efficiently Training \\Large-scale Neural Networks through Federated Learning}

\author{
  Yuanyuan Chen$^{1*}$
  \and
  Zichen Chen$^{2*}$
  \and
  Pengcheng Wu$^1$
  \And
  Han Yu$^1$
  \affiliations
  $^1$School of Computer Science and Engineering, Nanyang Technological University, Singapore\\
  $^2$University of California, Santa Barbara, CA, USA\\
  \emails
  \{yuanyuan.chen, pengchengwu, han.yu\}@ntu.edu.sg, zichen\_chen@ucsb.edu
}

\begin{document}

\maketitle

\begin{abstract}
  Large-scale neural networks possess considerable expressive power. They are well-suited for complex learning tasks in industrial applications. However, large-scale models pose significant challenges for training under the current Federated Learning (FL) paradigm. Existing approaches for efficient FL training often leverage model parameter dropout. However, manipulating individual model parameters is not only inefficient in meaningfully reducing the communication overhead when training large-scale FL models, but may also be detrimental to the scaling efforts and model performance as shown by recent research. To address these issues, we propose the~\emph{Federated Opportunistic Block Dropout} (\methodname) approach. The key novelty is that it decomposes large-scale models into semantic blocks so that FL participants can opportunistically upload quantized blocks, which are deemed to be significant towards training the model, to the FL server for aggregation. Extensive experiments evaluating~\methodname{} against four state-of-the-art approaches based on multiple real-world datasets show that it reduces the overall communication overhead by more than $88\%$ compared to the best performing baseline approach, while achieving the highest test accuracy. To the best of our knowledge,~\methodname{} is the first approach to perform dropout on FL models at the block level rather than at the individual parameter level.
\end{abstract}

\section{Introduction}
Over the years, machine learning techniques have been applied to solve a wide range of problems~\cite{shi2023transferable,luo2023deep,shi2021fast,shi2022selective}. The size of the learning model matters. Both manually designed neural architectures and those generated through neural architecture search~\cite{elsken2019neural} demonstrate that scaling up model sizes can improve performance. However, it is a significant challenge to train a large-scale model, especially in a distributed manner (e.g., involving multiple collaborative organizations~\cite{Yu-et-al:2017}).
Federated Learning (FL)~\cite{yang2019federated} is a distributed machine learning paradigm that enables multiple data owners to collaboratively train models (e.g., neural networks (NNs)) without exposing their private sensitive data. In the centralized FL architecture, local model updates are aggregated into a global model in a privacy-preserving manner by a central parameter server~\cite{konevcny2016federated}. FL has been applied in many commercial scenarios such as enhancing user experience~\cite{yang2018applied}, safety monitoring~\cite{liu2020fedvision,Xie-et-al:2022} and healthcare~\cite{Liu-et-al2022IAAI}.

However, training large-scale deep models in current FL settings is challenging. The communication overhead involved is high~\cite{kairouz2019advances}. Typical FL approaches require local model updates to be sent to the server and the aggregated global model to be distributed to the clients multiple times. This limits the scale of neural networks that can be used, making them less well-suited for solving complex real-world problems. When multiple institutional data owners collaboratively train large-scale FL models, communication efficient FL training approaches are required.
\def\thefootnote{\arabic{footnote}}

One popular approach for achieving communication efficient training of large-scale NNs in FL is through parameter dropout.
FedDropoutAvg~\cite{FedDropoutAvg} randomly selected a subset of model parameters to be dropped out to reduce the size of transmission. In addition, it also randomly dropped out a subset of FL clients to further reduce the number of messages required. In contrast, Adaptive Federated Dropout (AFD)~\cite{fed-dropout} performed dropout based on parameter importance by maintaining an activation score map. In such approaches, it is common to apply compression techniques to further reduce the communication overhead when transmitting model updates to and from the FL server.\footnote{Note that although Federated Dropout~\cite{fed_dropout}, FjORD~\cite{FjORD} and FedDrop~\cite{fed_drop} also leveraged parameter dropout in FL, their design goal was to enhance model adaptability to FL client heterogeneity, rather than training large-scale NN models efficiently via FL.}
However, these approaches directly manipulate individual model parameters. Recent research has shown that this is not only inefficient in terms of meaningfully reducing the communication overhead when training large-scale FL models, but may also negatively affect the scaling efforts and model performance~\cite{Gary-et-al:2022}.

To address this limitation, we propose the \emph{Federated Opportunistic Block Dropout} (\methodname) approach. By dividing large-scale deep models into semantic blocks, it evaluates block importance (instead of determining individual parameter importance) and opportunistically discards unimportant blocks in order to enable more significant reduction of communication overhead while preserving model performance. Since the block importance measure is not based on the client loss function as is the case for AFD~\cite{fed-dropout}, \methodname{} can handle complex tasks effectively.

We study the performance of \methodname{} for training large-scale deep FL models through extensive experiments in comparison with four state-of-the-art baseline approaches on three real-world datasets (including CIFAR-10, CIFAR-100 and IMDB). The results show that, compared to the best performing existing approach, \methodname{} reduces the communication overhead by $88\%$, while achieving the highest test accuracy.
%It achieves the most advantageous trade-off between communication overhead and test accuracy among all comparison approaches.
To the best of our knowledge,~\methodname{} is the first semantic block importance-based opportunistic FL dropout approach.

\section{Related Work}

Existing methods for improving FL communication efficiency can be divided into two major categories:

\paragraph{Compression.}
Deep Gradient Compression (DGC)~\cite{lin2017deep} employed gradient sparsification to reduce redundant gradients, and thereby enabling large models to be trained more efficiently under FL settings. FetchSGD~\cite{rothchild2020fetchsgd} took advantage of Count Sketch to compress gradient updates, while accounting for momentum and error accumulation in the central FL server. Nevertheless, it required a large number of communication rounds to achieve convergence because aggregation is performed after each local batch. SignSGD~\cite{bernstein2018signsgd} was a typical method for compressing model gradients. However, the compression was static which can result in loss of important features and convergence to a global model with degraded performance.

In recent years, approaches that directly compress FL model parameters are starting to emerge~\cite{amiri2020federated,reisizadeh2020fedpaq}. This is more challenging compared to gradients compression as more training information is lost in the process. Nevertheless, these approaches can reduce the number of communication rounds required during FL model training, and can achieve comparable performance with full-precision networks~\cite{hou2018analysis}. For example, FedPAQ~\cite{reisizadeh2020fedpaq} compressed the model updates before sending them to the FL server. However, it required a static quantization approach to be applied before the compression step. Thus, it was only able to support simple learning tasks and small-scale neural networks. Existing compression methods are not suitable for complex FL tasks involving large-scale neural networks.

\paragraph{Parameter dropout.}
Dropout methods for FL training have been proposed to address two distinct problems: 1) enabling devices with diverse local resources to collaboratively train a model, and 2) enabling efficient training of a large-scale FL model by organizational data owners (which are not resource constrained).

Federated Dropout~\cite{fed_dropout} exploited user-server model asymmetry to leverage the diverse computation and communication capabilities possessed by FL clients to train a model which could be too large for a subset of the clients to handle. It fixed the server model size and applied parameter dropout at different rates to generate models suitable for each client to train according to their local resource constraints.
FjORD~\cite{FjORD} extended Federated Droput to propose the ordered dropout method, under which the client sub-models were selected in a nested fashion from the FL server model. In this way, each client was assigned a model of size in proportion to its local computational resources for training, thereby adapting the model according to client device heterogeneity. Similarly, FedDrop~\cite{fed_drop} incorporated additional dropout layers into FL models to determine the channel wise trajectories to be dropped or retained to adapt a model according to local data distributions of FL clients.
Nevertheless, these approaches were designed to address the first problem, which is not the focus of this paper.

Our research focuses on the second problem. The FedDropoutAvg approach~\cite{FedDropoutAvg} randomly dropped out a subset of model parameters while randomly dropping out some FL clients before performing FedAvg based model aggregation. The Adaptive Federated Dropout (AFD) approach~\cite{fed-dropout} adaptively determined a percentage of weights based on parameter importance to be dropped. In this way, compression can be performed to reduce the communication overhead of transmitting the model updates to the FL server. It maintained an activation score map which is used to determine the importance of the activations to the training process, and determine which of them shall be kept or dropped in the current round of training. Instead of determining individual parameter importance, the proposed~\methodname{} approach focuses on evaluating the importance of semantic blocks in a large-scale deep model and opportunistically discarding unimportant blocks in order to enable more significant reduction of communication overhead while preserving model performance, thereby enabling efficient training of large-scale FL models by institutional data owners.

\section{The Proposed \methodname{} Approach}
\begin{figure*}[!t]
  \centering
  \includegraphics[width=0.9\textwidth]{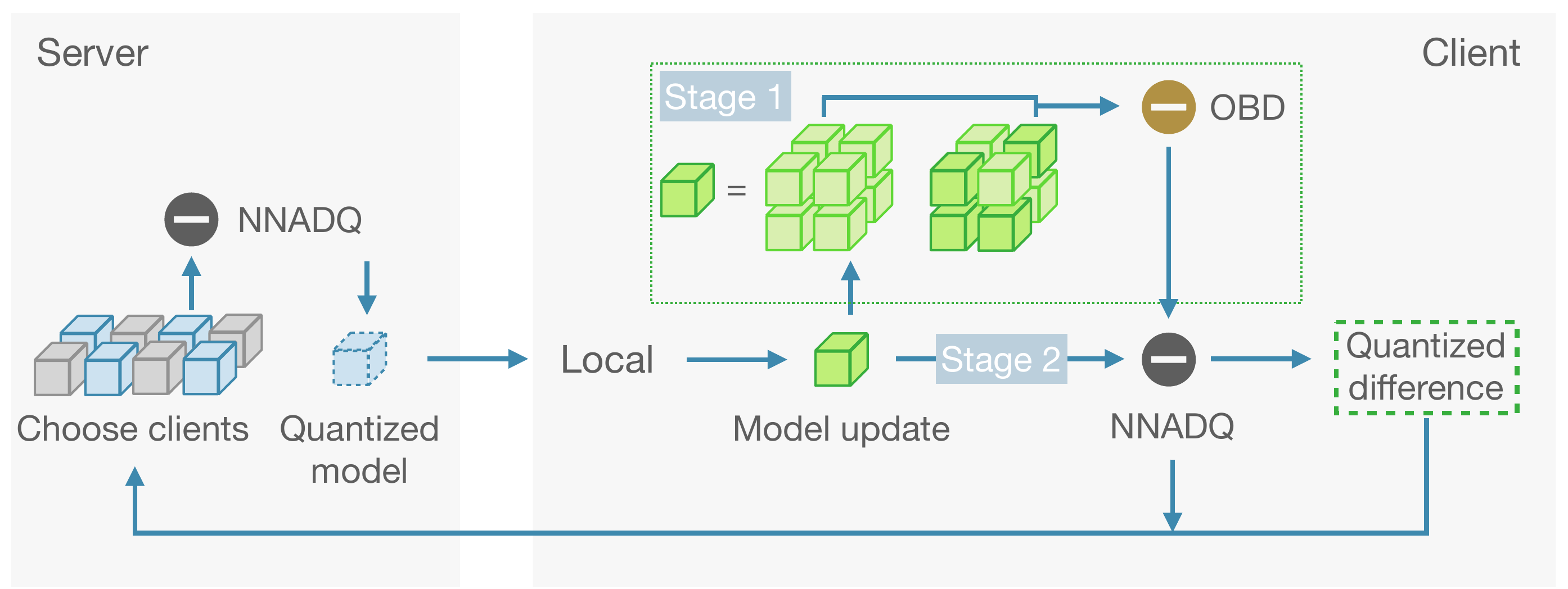}
  \caption{An overview of~\methodname{}} \label{fig:FedOBD}
\end{figure*}

FL is a distributed machine learning paradigm involving multiple data owners (a.k.a., FL clients) to collaboratively train models under the coordination of a parameter server (a.k.a., FL server).\footnote{Here, we refer to the horizontal federated learning setting with an FL server~\cite{yang2019federated}.}
Formally, assume that there are $n$ clients participating in FL model training. Each client $i$ owns a local dataset $D_i = \left \{  \left (  \vect{x}_j, \vect{y}_j \right )  \right \}_{j=1}^{M_i}$, where $\vect{x}_j$ is the $j$-th local training sample, $\vect{y}_j$ is the corresponding ground truth label and $M_i$ is the number of samples in $D_i$.
Under these settings, FL aims to solve the following optimization problem:
\begin{equation}
  \min_{\param\in \hypothesisspace}\sum_{i=1}^{n}\frac{M_i}{M} \mathcal{L}_{i}(\param;D_i) ,
\end{equation}
where $\hypothesisspace$ is the parameter space determined by the neural network architecture. $M \defeq \Sigma_{i=1}^n M_i$ is the total number of samples and $\mathcal{L}_{i}(\param;D_i) \defeq \frac{1}{M_i} \sum_{j=1}^{M_i} \ell(\param;\vect{x}_j, \vect{y}_j)$ is the local loss of client $i$.
Normally, the training process consists of multiple synchronization rounds. In the beginning of each round, the FL server distributes a global model to selected FL clients. The clients train the received models using local data and send the resulting model updates back to the server. The server then combines these uploaded models into an updated global model following some aggregation algorithm (e.g., FedAvg~\cite{pmlr-v54-mcmahan17a}). These steps are repeated until model convergence.
\methodname{} is designed for the aforementioned horizontal Federated Learning setting. It attempts to identify unnecessary model updates during training to reduce communication overhead while maintaining model performance. The following subsections explain the key components of~\methodname{} to achieve this design goal. They are also illustrated in figure~\ref{fig:FedOBD}.

\subsection{Opportunistic Block Dropout (OBD)}
Neural networks are often designed with some structures in mind. Their generalization performance relies heavily on properly selecting the structures. Motivated by this observation, a neural network can be decomposed into blocks of consecutive layers before FL model training commences. The key idea of~\methodname{} is to identify important blocks and only involve them, instead of the entire model, during the FL training process in order to reduce communication overhead. To achieve this goal, we need to address three questions: 1) how to decompose a given model; 2) how to determine block importance; and 3) how to aggregate the uploaded blocks.

When decomposing the trained model, it is critical to recognize popular (i.e., frequently used) neural network structural patterns. For example, layer sequences such as $\langle$Convolution, Pooling, Normalization, Activation$\rangle$ are commonly found in convolutional neural networks, and Encoder layers are commonly found in Transformer based models. Other types of architecture may define basic building blocks which can be taken into consideration when dividing a model into blocks. Such blocks often provide important functionality (e.g., feature extraction). Thus, it makes sense to transmit or drop them as a whole. Finally, the remaining layers are treated as singleton blocks.

\begin{algorithm}[t!]
  \caption{OBD}
  \label{algo:block}
  \SetKwInOut{KwIn}{Input}
  \SetKwInOut{KwOut}{Output}
  \SetKw{Continue}{continue}
  \KwIn{global model $\param_{r-1}$, local model $\param_{r,i}$ in client $i$, the set of identified block structures $B$, dropout rate $\dropoutrate\in [0,1]$. }
  \KwOut{retained blocks.}

  ${important\_blocks} \gets\codename{MaxHeap}()$;

  \ForEach{$\vect{b} \in B$}
  {
    ${important\_blocks}.\codename{push}(\vect{b}_{r,i})$ with key $\funname{MBD}(\vect{b}_{r-1}, \vect{b}_{r,i}) $;
  }
  ${revised\_model\_size} \gets 0$; \\
  ${retained\_blocks} \gets \codename{List}()$; \\
  \While{${important\_blocks}$ is not empty}
  {
    $\vect{b}_{r,i} \gets {important\_blocks}.\codename{pop}()$;  \\
    ${new\_size} \gets {revised\_model\_size} + \abs{\codename{asVect}(\vect{b}_{r,i})}$; \\
    \If{ ${new\_size} > (1-\dropoutrate)\abs{\emph{\codename{asVect}}(\param_{r,i})}$ }
    {
      \Continue{}; \\
    }
    ${revised\_model\_size} \gets  {new\_size}$; \\
    ${retained\_blocks}.\codename{append}(\vect{b}_{r,i})$;
  }
  \Return{${retained\_blocks}$};
\end{algorithm}

The proposed opportunistic block dropout approach is shown in algorithm~\ref{algo:block}. At the end of each round of local training, each block is assigned an importance score. To achieve this goal, a client $i$ keeps a copy of the global FL model $\param_{r-1}$ and compares it with the resulting model $\param_{r,i}$ block by block after local training. We propose the Mean Block Difference (MBD) metric to measure block importance, defined as
\begin{equation}
  \funname{MBD}(\vect{b}_{r-1}, \vect{b}_{r,i}) \defeq \frac{\norm{\codename{asVect}(\vect{b}_{r-1}) - \codename{asVect}(\vect{b}_{r,i})}}{\abs{\vect{b}_{r-1}}},
\end{equation}
where $\vect{b}_{r-1}$ denotes the blocks in the received global FL model, and $\vect{b}_{r,i}$ denotes the corresponding blocks in the local model produced by client $i$ after the current round of training. The $\codename{asVect}$ operator flattens tensors into a single vector and $\abs{\vect{b}}$ gives the number of parameters in a block.
In general, the larger MBD value between the new and old versions of the same block, the more important the new block is because it may contain lots of new knowledge learned in the current round.

Once the MBD values for all the blocks have been computed,~\methodname{} determines which blocks to retain (lines 5-15). Since all the blocks are ranked from high to low in terms of their MBD values with a max-heap, the algorithm simply pops each block one by one in descending order of their MBD values and adds each of them to the ${retained\_blocks}$ list, while keeping track of the size of the revised model so far.
When the algorithm terminates, the size of the revised model is guaranteed to not exceed $(1-\dropoutrate)\abs{\codename{asVect}(\param_{r,i})}$, where $\dropoutrate\in[0,1]$ is the dropout rate (where 1 indicates dropping out the entire model and 0 indicates no dropout). Only the retained blocks are transmitted to the FL server for model aggregation.
% As soon as the size of the revised model exceeds $(1-\dropoutrate)\lonenorm{\funname{vector}(\param_{r,i})}$, where $\dropoutrate\in[0,1]$ is the dropout rate (where 1 indicates no output and 0 indicates no dropout), the algorithm terminates and only the retained blocks are transmitted to the FL server for model aggregation.
% Note that here, each retained block is stored in the form of the differences between the corresponding parameter values in $\vect{b}_{r,i}$ and $\vect{b}_{r-1}$.

%Intuitively, those blocks that ``learned'' most knowledge during local training are chosen while others are dropped. Furthermore, block differences are computed and uploaded rather than blocks to minimize the chance of model leakage.

Upon receiving retained blocks from a client $i$ in round $r$, the FL server combines them with unchanged blocks from the previous global model $\param_{r-1}$ to form $\param_{r,i}$. The reconstructed local models for all clients are then used to aggregate a new global FL model $\param_r$ following an aggregation algorithm such as FedAvg~\cite{pmlr-v54-mcmahan17a}.

\begin{algorithm}[t!]
  \caption{NNADQ}\label{al:NNADQ}
  \SetKwInOut{KwIn}{Input}
  \KwIn{layer-structured tensors $tensors$, relative weight $\relativeweight$.}
  \KwOut{quantized tensors and other parameters for dequantization.}

  $ results \gets \codename{List}()$; \\

  \ForEach{tensor $\vect{t} \in tensors$}
  {
    $ \vect{v} \gets \codename{asVect}(\vect{t})$; \\
    $ result \gets \codename{ADQ}(\vect{v}, \relativeweight)$;   \\
    change quantized $\vect{v}$ in $result$ back to tensor form; \\
    $results.\codename{add}(result)$;   \\
  }
  \Return{$results$}; \\
\end{algorithm}

\begin{algorithm}[t!]
  \caption{ADQ}\label{al:ADQ}
  \SetKwInOut{KwIn}{Input}
  \SetKwInOut{KwOut}{Output}
  \KwIn{vector $\vect{v}$ to quantize, relative weight $\relativeweight$.}
  \KwOut{a quantized vector and parameters for dequantization.}
  $ {max\_v}, {min\_v} \gets \funname{maxmin}(\vect{v})$; \\
  $ offset \gets \argmin_\theta~\max({max\_v}+\theta, {min\_v}+\theta)$; \\
  $\vect{v}^\prime \gets \vect{v}+offset$; \\
  $\normvar \gets  \infinitynorm{\vect{v}^\prime}$; \\
  $\vect{sgn} \gets \funname{sgn}{( \vect{v}^\prime)}$; \\
  $\quantlevelnumber \gets \lfloor \max \left(\sqrt{\ln{4}*\frac{REPR}{\relativeweight}*d},1\right) \rfloor$; \\
  $\vect{v}_{\text{quant}} \gets \funname{round}(\vect{v}^\prime, {\quantlevelnumber}, \normvar)$; \\
  \Return{$\langle\vect{v}_{\text{quant}},\vect{sgn}, {offset}, \normvar, {\quantlevelnumber}\rangle$}; \\
\end{algorithm}

\begin{algorithm}[t!]
  \SetKwInOut{KwIn}{Input}
  \SetKwInOut{KwOut}{Output}
  \KwIn{number of FL clients $n$, \\
    client subset size $k$, \\
    dropout rate $\dropoutrate\in [0,1]$,\\
    relative weight $\relativeweight$ in quantization, \\
    number of rounds $R$ in stage 1, \\
    first stage number of local epochs $E_1$,\\
    second stage number of local epochs $E_2$.
  }
  \KwOut{the final global FL model.}

  \textbf{Stage 1:}\\
  //\quad{}at the FL server:\\
  initialize $\param_0$;\\
  \ForEach{$r\in\{1,\ldots,R\}$}
  {
  ${C}_r \gets$ $k$ randomly chosen clients; \\
  distribute $\codename{NNADQ}(\param[r-1],\relativeweight)$ to ${C}_r$; \\

  //\quad{}at each client $i \in\ {C}_r$:\\
  {
  dequantize data, construct and load $\param_{r-1}$; \\
  train $\param_{r,i}$ for $E_1$ epochs; \\
  ${important\_blocks} \gets \codename{OBD}(\param_{r,i}, \param_{r-1}, \dropoutrate)$; \\
  upload $\codename{NNADQ}(\codename{diff}({important\_blocks}, {old\_blocks}), \relativeweight$) to FL server; \\
    }
    //\quad{}at the FL server:\\
    \ForEach{client $i \in {C}_r$}
    {
      dequantize the received update and reconstruct $\param_{r,i}$; \\
    }
  $\param_{r} \gets \codename{aggregate}(\{\param_{r,i}\}_{i \in {C}_r})$ \\
    }
    \textbf{Stage 2:}\\
    %each client to set local epochs to $E_2$;\\
    //\quad{}at the FL server:\\
  $\param_0 \gets \param_{R}$ \\
    \ForEach{epoch $e\in\{1,\ldots,E_2\}$}
    {
    distribute $\codename{NNADQ}(\param_{e-1}, \relativeweight)$ to clients; \\
    //\quad{}at each client $i\in\{1,\ldots,n\}$:\\
    {
    dequantize data, construct and load $\param_{e-1}$; \\
  $\param_{e,i} \gets $ train with previous learning rate for 1 epoch; \\
    upload $\codename{NNADQ}(\codename{diff}(\param_{e,i},\param_{e-1}), \relativeweight)$ to FL server; \\
    }
    //\quad{}at the FL server:\\
    \ForEach{$i\in\{1,\ldots,n\}$}
    {
      dequantize the received update and reconstruct $\param_{e,i}$; \\
    }
  $\param_e \gets \codename{aggregate}(\param_{e,1}, \ldots{}, \param_{e,n})$; \\
  }
  \caption{\methodname{}}
  \label{algo:fedobd}
\end{algorithm}

\subsection{Adaptive Deterministic Quantization for Neural Networks}
\label{section:NNADQ}
While OBD reduces the transmission size of the original model by retaining only important blocks based on a user specified dropout rate, to further reduce communication overhead, we propose an improved quantization approach. In~\methodname{}, local block differences and global models are quantized before being sent out. The following parts describe our NNADQ quantization approach in detail.

\subsubsection*{Stochastic Quantization}
Quantization reduces the number of bits to represent a vector element by converting it to an element of a finite number of quantization levels $\mathcal{Q}$. Therefore, quantization can be roughly described by a function $\funname{quant}: \realset^n \rightarrow \mathcal{Q}^n$. Some quantization functions are paired with a dequantization function $\funname{dequant}: \mathcal{Q}^n \rightarrow \realset^n $ to recover the original data approximately.

Our quantization approach is inspired by~\emph{Stochastic Quantization} proposed in~\cite{alistarh2017qsgd}. Given the number of quantization levels $\quantlevelnumber$, Stochastic Quantization is formulated as
\begin{equation*}
  \begin{split}
    \funname{quant}_\text{SQ}(\vect{v},\quantlevelnumber) & \defeq (\norm{\vect{v}}\quantlevelnumber, \sgn{(\vect{v})}, \xi(\vect{v},\quantlevelnumber)), \\
    \funname{dequant}_\text{SQ}(n,\vect{sgn},\vect{v})    & \defeq n*(\vect{sgn} \circ \vect{v}) ,                                                        \\
  \end{split}
\end{equation*}
where $\xi(\vect{v},\quantlevelnumber)$ is a random vector explained below, $\circ$ is the element-wise multiplication and $*$ is the scalar-vector multiplication. The quantization levels are natural numbers $\{0,\dots,\quantlevelnumber-1\}$.

Let $0 \leq l_i < \quantlevelnumber$ be the unique integer such that $\abs{v_i}/\norm{\vect{v}} \in [ l_i / \quantlevelnumber, (l_i + 1) / \quantlevelnumber ]$, that is, $[l_i / \quantlevelnumber, (l_i + 1) / \quantlevelnumber]$ is a quantization interval for $\abs{v_i}/\norm{\vect{v}}$.
Then,
\begin{equation}
  \label{eq:SQrandom}
  \xi(\vect{v},\quantlevelnumber)  \defeq
  \begin{cases}
    l_i+1 & \mbox{with probability $(\frac{\abs{v_i}}{\norm{\vect{v}}}\quantlevelnumber-l_i)$}, \\
    l_i   & \mbox{otherwise}.                                                                   \\
  \end{cases}
\end{equation}
It can be proven that $\funname{quant}_\text{SQ}(\vect{v},\quantlevelnumber)$ is an unbiased estimation of $\vect{v}$.

\subsubsection*{Reformulation}
Although Stochastic Quantization gives a statistically unbiased output, it is still unclear whether the form of normalization is optimal. It is also necessary to reformulate the problem taking the balance between compression and informativeness into consideration.

Following the same notation, if $\quantlevelnumber$ and $\normvar$ are treated as variables, equation~\eqref{eq:SQrandom} can be rewritten as
\begin{equation*}
  \zeta(\vect{v},\quantlevelnumber,\normvar) \defeq \left\{ \begin{array}{ll}
    l_i+1 & \mbox{with probability $(\frac{\abs{v_i}}{\normvar}\quantlevelnumber-l_i)$}, \\
    l_i   & \mbox{otherwise}.\end{array} \right.
\end{equation*}

Since $\zeta/\quantlevelnumber$ is an unbiased estimator of $\vect{v}/\normvar$, $\sum_{i}{}\funname{Var}(\zeta_i/\quantlevelnumber)$ is an indication of quantization loss. However, we quickly find that
\begin{equation*}
  \funname{round}(\vect{v},\quantlevelnumber,\normvar) \defeq
  \begin{cases}
    l_i   & \mbox{if $\frac{\abs{v_i}}{\normvar}$ is closer to $ \frac{l_i}{\quantlevelnumber}$}, \\
    l_i+1 & \mbox{otherwise},                                                                     \\
  \end{cases}
\end{equation*}
is a better alternative in terms of quantization loss since $ \sum_{i}{}(\abs{v_i}/d-\funname{round}_i(\vect{v},\quantlevelnumber,\normvar)/\quantlevelnumber)^2 \leqslant \sum_{i}{}\funname{Var}(\zeta_i/\quantlevelnumber) $, and the new way of quantization works deterministically rather than stochastically.

The trade-off between compression and informativeness can now be formulated as an optimization problem:
\begin{equation}
  \label{problem:tradeoff}
  \begin{split}
    \min_{\quantlevelnumber,\normvar} \quad & \frac{d^2}{n}\sum_i{ \left(\frac{\abs{v_i}}{\normvar}-\frac{\funname{round}_i(\vect{v},\quantlevelnumber,\normvar)}{\quantlevelnumber}\right)^2 } \\
                                            & +\relativeweight\frac{
      \lceil\log_2{(\quantlevelnumber+1)} \rceil
    }
    {REPR},                                                                                                                                                                                     \\
    \subjectto \quad                        & \quantlevelnumber \geqslant 1, \quantlevelnumber \in \naturalnumberset,                                                                           \\
    \quad                                   & d \geqslant \infinitynorm{\vect{v}}.                                                                                                              \\
  \end{split}
\end{equation}
The first part of equation~\eqref{problem:tradeoff} indicates the mean information loss and the second part indicates the compression ratio where $REPR$ is the number of bits in a floating-point representation\footnote{$REPR=32$ in typical platforms.}. $\relativeweight \in \realset^{+}$ is a predefined relative weight between the two parts. \\
Since $\quantlevelnumber{}l \leqslant \quantlevelnumber\abs{v_i} \leqslant \quantlevelnumber{}l+\normvar$, by definition, we have
\begin{equation}
  \label{eq:upper_bound}
  \begin{aligned}
     & \frac{\normvar^2}{n}\sum_i{\left(\frac{\abs{v_i}}{\normvar}-\frac{\funname{round}_i(\vect{v},\quantlevelnumber,\normvar)}{\quantlevelnumber}\right)^2 }                           \\
     & = \frac{\normvar^2}{n}\sum_i{ \left(\min \left(\frac{\abs{v_i}}{\normvar}-\frac{l}{\quantlevelnumber}, \frac{\abs{v_i}}{\normvar}-\frac{l+1}{\quantlevelnumber}\right)\right)^2 } \\
     & \leqslant \frac{\normvar^2}{n}\sum_i{\left(\frac{\abs{v_i}}{\normvar}-\frac{l}{\quantlevelnumber}\right)\left(\frac{\abs{v_i}}{\normvar}-\frac{l+1}{\quantlevelnumber}\right) }   \\
     & =\frac{1}{n\quantlevelnumber^2}\sum_i{ (\quantlevelnumber\abs{v_i}-\quantlevelnumber{}l)(\normvar+\quantlevelnumber{}l-\quantlevelnumber\abs{v_i}) }                              \\
     & \leqslant \frac{1}{n\quantlevelnumber^2} \sum_i{(\normvar+\quantlevelnumber{}l-\quantlevelnumber\abs{v_i})} \leqslant \frac{\normvar}{\quantlevelnumber^2}.
  \end{aligned}
\end{equation}

Equation~\eqref{eq:upper_bound} shows that $\normvar$ should be as small as possible and $\normvar=\infinitynorm{\vect{v}}$ is a near-optimal choice for any fixed $\quantlevelnumber$.
For any fixed $\normvar$, we allow $\quantlevelnumber$ to take real values and combine equation~\eqref{problem:tradeoff} and equation~\eqref{eq:upper_bound} to obtain a simplified optimization objective
\begin{equation}
  \label{eq:objective_real}
  \begin{aligned}
    \min_{\quantlevelnumber} \quad & \frac{\normvar}{\quantlevelnumber^2}+\relativeweight \frac{\log_2{\quantlevelnumber}+1}{REPR}, \\
    \subjectto \quad               & \quantlevelnumber \geqslant 1, \quantlevelnumber \in \realset.                                 \\
  \end{aligned}
\end{equation}

Solving equation~\eqref{eq:objective_real} gives $\quantlevelnumber^*_{\realset}=\max \left(\sqrt{\ln{4}*\frac{REPR}{\relativeweight}*\normvar}, 1\right)$. $ \lfloor \quantlevelnumber^*_{\realset}  \rfloor$ is then taken as an approximate solution to the original problem equation~\eqref{problem:tradeoff}.

\subsubsection*{Further Optimization}
Note that $\normvar=\infinitynorm{\vect{v}}$ affects the information loss directly and the compression ratio indirectly via $\quantlevelnumber^*_{\realset}$. Smaller values of $\infinitynorm{\vect{v}}$ can improve equation~\eqref{problem:tradeoff} in general. Thus, we should translate $\vect{v}$ to $\vect{v}^\prime$ with minimum infinity norm before quantizing it. Since each layer of a large-scale model has a different statistical parameter distribution, quantization layer by layer can utilize this trick further.

Based on the aforementioned quantization approach, we propose the Adaptive Deterministic Quantization for Neural Networks (NNADQ) approach as in algorithm~\ref{al:NNADQ} (and the related supporting function in algorithm~\ref{al:ADQ}). NNADQ is used in~\methodname{} to compress both directions of transmission.

\subsection{The Two-Stage Training Process}
Finally,~\methodname{} has a two-stage training process.
In the first stage, small local epochs are used, random subsets of clients are selected in each round and OBD is activated (i.e., only selected blocks are uploaded to the server). In this way,~\methodname{} encourages frequent aggregation to prevent overfitting without incurring high communication overhead.

In the second stage,~\methodname{} switches to a single round with aggregation being executed at the end of each epoch so that local learning rates are reused. Therefore,~\methodname{} attempts to fine-tune the global model by approximating centralized training.
The detailed process of~\methodname{} is presented in algorithm~\ref{algo:fedobd}.

\section{Experimental Evaluation}

\begin{table*}[!t]
  \centering
  \caption{Performance and communication efficiency of various FL approaches.} \label{table:performance}
  \begin{tabular}{lccc}
    \toprule
    {Dataset}                  & {Approach}    & {Data transmission (MB)}          & {Test accuracy}             \\
    \midrule
    \multirow{5}{*}{CIFAR-10}  & FedAvg        & \num{13504.22}                    & $\mathbf{82.75 \pm 2.53\%}$ \\
                               & FedPAQ        & \num{4266.26}                     & $81.75 \pm 0.47\%$          \\
                               & FedDropoutAvg & $\num{5777.91} \pm 0.05$          & $81.59 \pm 0.38\%$          \\
                               & SMAFD         & $\num{2495.13} \pm 18.57$         & $27.62 \pm 12.34\%$         \\
                               & \methodname{} & $\mathbf{1440.96\pm 4.95}$        & $82.70 \pm 0.28\%$          \\
    \vspace{.5\baselineskip}                                                                                     \\
    \multirow{5}{*}{CIFAR-100} & FedAvg        & \num{7211.02}                     & $53.06 \pm 0.48\%$          \\
                               & FedPAQ        & \num{2278.10}                     & $52.67 \pm 0.58\%$          \\
                               & FedDropoutAvg & $\num{3085.31}\pm 0.06$           & $52.62 \pm 0.42\%$          \\
                               & SMAFD         & $\num{1367.12} \pm 10.14$         & $9.27 \pm 7.07\%$           \\
                               & \methodname{} & $\mathbf{811.79\pm 1.57}$         & $\mathbf{53.11 \pm 0.62\%}$ \\
    \vspace{.5\baselineskip}                                                                                     \\
    \multirow{5}{*}{IMDB}      & FedAvg        & \num{1321669.23}                  & $77.78\pm 1.36\%$           \\
                               & FedPAQ        & \num{417542.27}                   & $78.69  \pm 1.41\%$         \\
                               & FedDropoutAvg & $\num{565490.25}\pm 0.58$         & $78.51 \pm 2.11\%$          \\
                               & SMAFD         & $\num{321706.16}\pm 3,039.03$     & $56.33 \pm 6.40\%$          \\
                               & \methodname{} & $\mathbf{178,359.36 \pm 1063.37}$ & $\mathbf{79.66\pm 1.22\%}$  \\
  \end{tabular}
\end{table*}

\begin{table*}[!t]
  \caption{Performance and communication efficiency of \methodname{} variants on CIFAR-100.} \label{table:variant}
  \centering
  \begin{tabular}{lcc}
    \toprule
    Variant                         & Data transmission (MB) & Test accuracy      \\
    \midrule
    \methodname{} with SQ           & 977.61                 & $52.89 \pm 0.68\%$ \\
    \methodname{} w/o 2nd Stage     & $673.88 \pm 1.07$      & $50.06 \pm 0.54\%$ \\
    \methodname{} w/o Block Dropout & $926.23 \pm 1.81$      & $55.85 \pm 1.08\%$ \\
  \end{tabular}
\end{table*}

\label{experiment}
In the experiments, the performance of~\methodname{} was compared against four state-of-the-art related approaches. Ablation studies were also conducted to investigate the impact of different components of~\methodname{} on its effectiveness.

\subsection{Experiment Settings}
We compared~\methodname{} with these approaches:
\begin{description}
  \item[FedAvg~\cite{pmlr-v54-mcmahan17a}] A classic FL approach used as the baseline.
  \item[FedPAQ~\cite{reisizadeh2020fedpaq}] A compression-based communication efficient FL model training approach, which utilizes stochastic quantization~\cite{alistarh2017qsgd} and random client selection.
  \item[Adaptive Federated Dropout~\cite{fed-dropout}] An FL approach that optimizes both server-client communication costs and computation costs by allowing clients to train locally on a selected subset of the global model parameters. The Single-Model Adaptive Federated Dropout (SMAFD) variant was used.
  \item[FedDropoutAvg~\cite{FedDropoutAvg}] An FL approach that randomly drops out a subset of model parameters, while randomly dropping out some clients before performing FedAvg based model aggregation.
\end{description}

\paragraph{Hardware \& software.} The experiments were carried out on a server with 4 NVIDIA A100 GPUs, 1 AMD EPYC CPU and 252 GB of memory. The algorithm and experiment implementations were based on the PyTorch~\cite{paszke2017automatic} framework.

\paragraph{Tasks.} Two image classification tasks on CIFAR-10, CIFAR-100~\cite{krizhevsky2009learning} and one sentiment classification task on IMDB~\cite{maas2011learning} were used.
CIFAR-10 consists of \num{50000} training images and \num{10000} testing images in 10 classes. CIFAR-100 consists of \num{50000} training images and \num{10000} testing images in 100 classes. IMDB consists of \num{25000} highly polar movie reviews for training, and \num{25000} reviews for testing in two classes. We considered only the i.i.d.~learning case: local training and validation datasets were drawn uniformly from the larger datasets and the server held a separate test dataset for evaluating.

\paragraph{Models.} For CIFAR-10 and CIFAR-100 tasks, two DenseNet-40 network variants were used, one with 0.17 million parameters and the other with 0.19 million parameters~\cite{huang2017densely}. A Transformer based classification network was used~\cite{he2016deep} in IMDB tasks. It consisted of 2 encoder layers followed by a linear layer, with a total of 17 million parameters. In addition, its word embedding layer was initialized with GloVe word embeddings~\cite{pennington2014glove}.

\paragraph{Experiment scale.} We investigated the performance of~\methodname{} and other approaches under large-scale learning settings. Specifically, 100 clients were used in CIFAR-10 and IMDB tasks and 50 clients were used in CIFAR-100 task. Furthermore, a fixed fraction of $50\%$ clients were chosen randomly in each round in FedPAQ, FedDropoutAvg, SMAFD and~\methodname{} since random client selection is part of these algorithms.

\paragraph{Independent trials.} To measure the stability of~\methodname{} and other approaches that use probabilistic methods, 10 independent trials were performed for each task.

\paragraph{Other hyperparameters.} 100 rounds and 5 local epochs were used in all combinations of algorithms and tasks for fair comparison. In addition,~\methodname{} used 10 epochs in the second stage.
Initial learning rates of 0.1 and 0.01 were used in image and sentiment classification tasks respectively and adjusted by Cosine Annealing learning rate strategy~\cite{loshchilov2016sgdr}. The batch size was set to 64. FedPAQ used 255 quantization levels. FedDropoutAvg, SMAFD and~\methodname{} used a dropout rate of 0.3.~\methodname{} used a quantization weight $\relativeweight$ of 0.001 in CIFAR-10/CIFAR-100 tasks and 0.0001 for IMDB.

\subsection{Results and Discussion}
Table~\ref{table:performance} lists mean and standard deviation values of performance and communication overheads over 10 independent trials for each combination of tasks and approaches. In each case, ``Data Transmission'' lists the amount of actual data transmission in megabytes and ``Test Accuracy'' evaluates the resulting model performance on the test dataset.

It can be easily calculated that, without parameter dropout/compression, FedAvg required around \num{20100} messages to be exchanged between the FL server and the clients, with $100\%$ of the model parameters being transmitted all the time in order to reach model convergence. Therefore, it achieved the highest test accuracies in large-scale training cases.
% SignSGD focuses on compressing the model gradient to be sent in each optimization step. It achieves the lowest transmission ratio among all the approaches consistently. However, in order for the model to converge under SignSGD, a significantly higher communication overhead is incurred compared to other approaches.
FedPAQ, FedDropoutAvg and SMAFD exchanged fewer messages due to parameter dropout. Overall, they achieved low communication overhead, with SMAFD consistently achieving the lowest overhead among them.

Compared to the above approaches,~\methodname{} required the lowest data transmission to achieve convergence. Notably, the communication overhead required by~\methodname{} was $88\%$ lower on average than that required by FedAvg. Meanwhile,~\methodname{} achieved the highest or second highest test accuracies, comparable to the performance of FedAvg. Overall,~\methodname{} achieved the most advantageous trade-off between communication overhead and test accuracy among all comparison approaches.

\subsection{Ablation Studies}
Since~\methodname{} consists of three key components, their effects need to be measured effect in isolation. Specifically, we considered the following variants of~\methodname{}:
\begin{description}
  \item[\methodname{} with SQ] In this variant, NNADQ is replaced with stochastic quantization~\cite{alistarh2017qsgd}.
  \item[\methodname{} w/o 2nd Stage] This variant only goes through the first stage of the training process.
  \item[\methodname{} w/o Block Dropout] The dropout rate $\dropoutrate$ is set to 0 in this variant.
\end{description}

Table~\ref{table:variant} lists mean and standard deviation values of performance and communication overheads over 10 independent trials for each variant under the CIFAR-100 task with the same hyperparameter settings as in the previous experiments.
Although ``\methodname{} with SQ'' achieved similar test accuracy compared to the canonical~\methodname{} method, it incurred $20.37\%$ higher communication overhead. This increment in communication overhead was a result of the transmission amount rising from $811.79$ MB under~\methodname{} to $977.15$ MB under ``\methodname{} with SQ''. This demonstrates the advantage of the proposed NNADQ approach in terms of reducing communication overhead.

When training without the second stage under ``\methodname{} w/o 2nd Stage'', an inferior global FL model was obtained which achieved $3\%$ lower test accuracy compared to the canonical~\methodname{} method. Nevertheless, without going through the second stage of training, it incurred $16.99\%$ lower communication overhead (as a result of less message exchanges) compared to the canonical~\methodname{} method. This shows that the proposed two-stage training scheme is necessary for producing high quality final models.

When trained under ``\methodname{} w/o Block Dropout'', the resulting model achieved higher test accuracy compared to the canonical~\methodname{} method (i.e., $2.7\%$ increase in test accuracy). However, it incurred $14.10\%$ higher communication overhead due to the transmission amount increasing from $811.79$ MB under~\methodname{} to $926.23$ MB. Hence, opportunistic block dropout is helpful in reducing communication overhead while preserving model performance.

Through the ablation studies, we have demonstrated that the algorithmic components of~\methodname{} are indeed indispensable towards achieving its design goal.

\section{Conclusion and Future Works}
In this paper, we set out to address an emerging research topic in the field of federated learning which is how to efficiently train large-scale deep models. This problem is commonly found in industrial FL application settings. We propose \methodname{}: a first-of-its-kind semantic block-level importance-based opportunistic dropout approach for improving FL model training efficiency, while maintaining model performance. Extensive experimental evaluation demonstrates that \methodname{} outperforms state-of-the-art baselines in terms of communication overhead and test accuracy.

We provide the reference implementation of~\methodname{} and related experiments in an open sourced project~\footnote{\url{https://github.com/cyyever/distributed_learning_simulator}.} for academic research. In the future, we plan to integrate more privacy-preserving considerations into the current~\methodname{} implementation to enable institutional data owners to collaboratively train complex large-scale FL models efficiently.

\section*{Acknowledgments}
This research/project is supported, in part, by the National Research Foundation Singapore and DSO National Laboratories under the AI Singapore Programme (AISG Award No: AISG2-RP-2020-019); Nanyang Technological University, Nanyang Assistant Professorship (NAP); the National Key R\&D Program of China (No. 2021YFF0900800); the Joint SDU-NTU Centre for AI Research (C-FAIR); and the RIE 2020 Advanced Manufacturing and Engineering (AME) Programmatic Fund (No. A20G8b0102), Singapore.

\section*{Contribution Statement}

The main contribution to this work is equally given by Yuanyuan Chen and Zichen Chen.

\bibliographystyle{named}
\bibliography{ijcai23}

\end{document}